\documentclass[final]{cvpr}

\usepackage{times}
\usepackage{epsfig}
\usepackage{graphicx}
\usepackage{amsmath}
\usepackage{amssymb}
\usepackage{multirow}
\usepackage{booktabs}
\usepackage{color}
\usepackage{xspace}
\usepackage{bm}

\usepackage[pagebackref=true,breaklinks=true,colorlinks,bookmarks=false]{hyperref}
\newcommand{\model}{M$^3$EM\xspace}
\newcommand{\modulesemantic}{SMR\xspace}
\newcommand{\modulecorr}{CMC\xspace}

\def\cvprPaperID{6754} 
\def\confYear{CVPR 2021}

\begin{document}

\title{EPIC-KITCHENS-100 Unsupervised Domain Adaptation Challenge for Action Recognition 2021: Team M$^3$EM Technical Report}

\author{Lijin Yang, Yifei Huang, Yusuke Sugano, Yoichi Sato\\
Institute of Industrial Science, the University of Tokyo\\
Tokyo, Japan\\
{\tt\small \{yang-lj,hyf,sugano,ysato\}@iis.u-tokyo.ac.jp}
}

\maketitle

\begin{abstract}
In this report, we describe the technical details of our submission to the 2021 EPIC-KITCHENS-100 Unsupervised Domain Adaptation Challenge for Action Recognition. Leveraging multiple modalities has been proved to benefit the Unsupervised Domain Adaptation (UDA) task. In this work, we present Multi-Modal Mutual Enhancement Module (\model), a deep module for jointly considering information from multiple modalities to find most transferable representations across domains. We achieve this by implementing two sub-modules for enhancing each modality using the context of other modalities. The first sub-module exchanges information across modalities through the semantic space, while the second sub-module finds the most transferable spatial region based on the consensus of all modalities.
\end{abstract}

\section{Introduction}
EPIC-KITCHENS-100 dataset contains fine-grained actions performed in different kitchens~\cite{damen2020rescaling}. How to make a model learned on a subset of kitchens (the source domain) to perform well on other unseen kitchens (the target domain) is challenging, since not only the verbs but also the associated objects can be dissimilar across domains. 

Previous works~\cite{munro2020multi,song2020modality} have shown that using multiple modalities can improve the performance of UDA on action recognition, but none of them considered using early fusion on the modalities to enhance the transferability of the generated features.
Intuitively, with the guidance of RGB, Flow can give more focus on the correct object, whereas by using knowledge from the motion, the RGB modality would concentrate more on the moving parts. In this paper, we argue that leveraging the information exchange across modalities before the final decision can significantly improve the transferability of features. Based on this intuitive, we propose a novel Multi-Modal Mutual Enhancement Module (\model) for domain adaptive action recognition by enhancing the features across modalities. 

The proposed \model consist of two sub-modules: a Semantic Mutual Refinement sub-module (\modulesemantic) and a Cross Modality Consensus sub-module (\modulecorr). To leverage the strength of each modality, the \modulesemantic enables information exchange across modalities through the semantic space. With \modulesemantic, a modality $M$ can receive recommendations from other modalities about transferable components that are easily ignored by the modality itself. The \modulecorr highlights the spatial regions that are transferable consistently in all the modalities. This sub-module complements \modulesemantic by preventing the \modulesemantic from emphasizing similar but irrelevant background regions that harm the action recognition. With the two proposed simple yet effective sub-modules, our module can be built on top of most existing domain adaptive action recognition models and improve their performance by integrating multi-modality signals.

\section{Method}
\begin{figure}
    \centering
    \includegraphics[width=\linewidth]{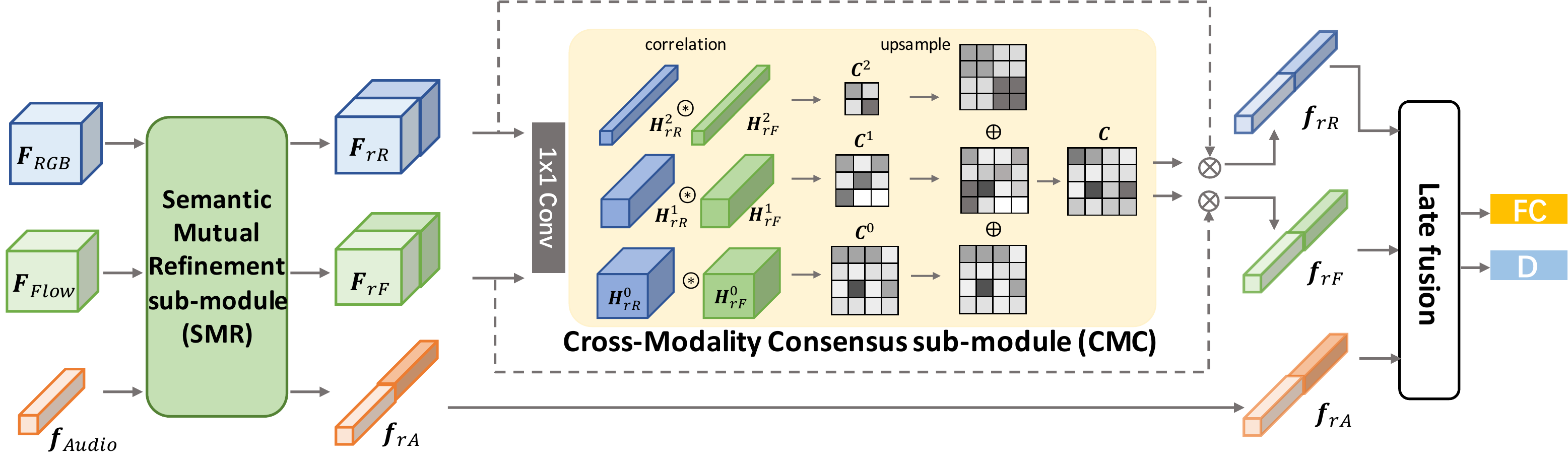}
    \caption{Overview of the proposed \model. We showcase three modalities RGB, Flow and Audio as input but it can be easily extended to add other modalities such as depth or hand. In the figure, $\oplus$ denotes element-wise summation, $\otimes$ is element-wise multiplication, and $\circledast$ means the correlation operation that calculates the Pearson correlation coefficient on each spatial position. \textit{FC} and \textit{D} are short for \textit{classifier} and \textit{discriminator}, respectively.}
    \label{fig:overview}
\end{figure}

Figure~\ref{fig:overview} depicts the overview of the proposed \model. For each modality of RGB, Flow and Audio, backbone (omitted in the figure) networks encode the input into frame-level features $\bm{F}_{RGB}, \bm{F}_{Flow}$ and $\bm{f}_{Audio}$, respectively. The features are then enhanced by information from other modalities using our proposed Semantic Mutual Refinement sub-module (\modulesemantic). For modality $M$, \modulesemantic summarizes information of $M$ and receive information from other modalities. Two gating functions are proposed to enhance the feature transferability by re-evaluating and re-mapping the transferable channels based on the summarized and received information. We then use a Cross-Modality Consensus sub-module (\modulecorr) to get the most transferable spatial region. \modulecorr finds the transferable region by calculating bit-wise correlation from different scales of the features. Finally, we adopt the adversarial learning framework by adding a discriminator to differentiate whether the input is from the source domain or not. We will introduce each component in detail in the following part of this section.

\subsection{The Semantic Mutual Refinement sub-module}
In this report, we propose a Semantic Mutual Refinement sub-module (\modulesemantic) to bridge the gap that prevents modality information exchange by channel re-evaluation and re-mapping. 

\begin{figure}[h]
    \centering
    \includegraphics[width=\linewidth]{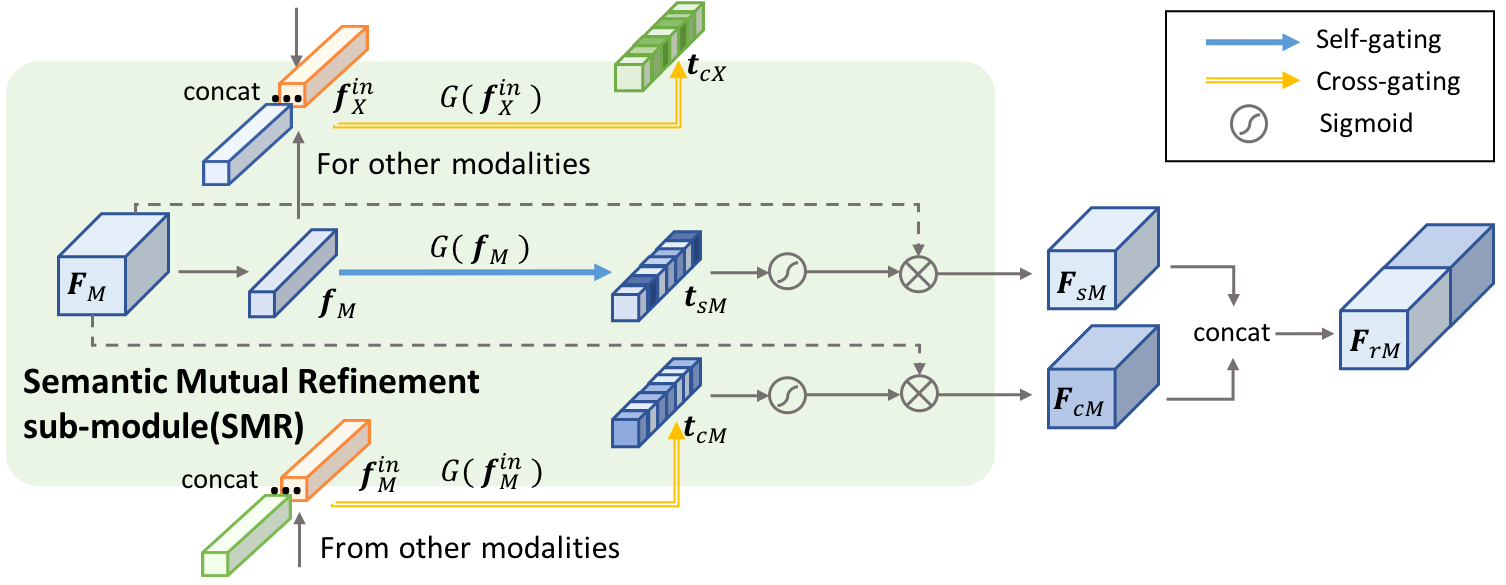}
    \caption{The Semantic Mutual Refinement sub-module (\modulesemantic) showcased using modality $M$. $M$ could be any modalities of RGB, Flow and Audio, also can be extended to other modalities if available, \eg, object.}
    \label{fig:smr}
\end{figure}

Figure~\ref{fig:smr} depicts the proposed \modulesemantic by showcasing the workflow of modality $M$. With \modulesemantic, the feature $\bm{F}_{M} \in \mathbb{R}^{c\times h\times w}$ encoded by the backbone will be enhanced to $\bm{F}_{rM} \in \mathbb{R}^{2c\times h\times w}$. $\bm{F}_{rM}$ is the concatenation of a self-refined feature $\bm{F}_{sM}$ and a cross-refined feature $\bm{F}_{cM}$. 
For getting $\bm{F}_{sM}$, a global average pooling is first conducted on $\bm{F}_{M}$ to obtain a global information embedding $\bm{f}_{M} \in \mathbb{R}^{c}$. 

$\bm{f}_{M}$ will re-evaluate the semantic transferability of modality $M$ itself by a self-gating function~\cite{hu2018squeeze}:
\begin{equation}
    \bm{t}_{sM} = G(\bm{f}_M) = \sigma \bm{W}^{M}_2(\delta (\bm{W}^{M}_1 \bm{f}_{M})),
\end{equation}
where $\bm{W}_1^M, \bm{W}_2^M$ are weight matrices, $\sigma$ and $\delta$ denotes the sigmoid and ReLU activations, respectively. Here $\bm{t}_{sM}$ is the re-evaluation of semantic transferability, and is used to emphasize the channels of $\bm{F}_{M}$ by element-wise multiplication on each spatial location $(i,j)$:
\begin{equation}
    \bm{F}_{sM}^{(i,j)} = \bm{F}_{M}^{(i,j)} \cdot \bm{t}_{sM},
\end{equation}

For getting $\bm{F}_{cM}$, we use a similar gating operation but with $\bm{f}^{in}_{M}$, the concatenation of features after global average pooling of all other modalities, as input. $\bm{f}^{in}_{M}$ serves as the recommendation information provided to modality $M$ by other modalities. We call this step cross-gating and represent it by:
\begin{equation}
    \bm{t}_{cM} = \sigma \bm{W}^{in}_2(\delta (\bm{W}^{in}_1 \bm{f}^{in}_{M})); \qquad
    \bm{F}_{cM}^{(i,j)} = \bm{F}_{M}^{(i,j)} \cdot \bm{t}_{cM},
\end{equation}
Thus, $\bm{F}_{cM}$ is the $M$ modality feature refined by other modalities via the cross-gating operation. 

It is important to prevent domain adaptation models from overfitting on the source domain. The \modulesemantic only introduces a small amount of model parameters by leveraging bottleneck during gating, \ie, we reduce the dimension by a ratio $r$ via making $\bm{W}_1 \in \mathbb{R}^{\frac{c}{r}\times c}$ and $\bm{W}_2 \in \mathbb{R}^{c \times \frac{c}{r}}$. Finally, we get the refined feature of modality $M$ by fusing the two refined features $\bm{F}_{sM}$ and $\bm{F}_{cM}$ via concatenation:
\begin{equation}
    \bm{F}_{rM} = Concat(\bm{F}_{sM}, \bm{F}_{cM}).
\end{equation}

\subsection{The Cross-Modality Consensus sub-module}
The structure of \modulecorr is shown in Figure~\ref{fig:overview}. This module first uses a 1x1 convolution layer on $\bm{F}_{rR}$ and $\bm{F}_{rF}$ for mapping the two modalities into a same latent space, formulating two features $\bm{H}_{rR}$ and $\bm{H}_{rF}$. Since the transferable regions vary in size in different samples, we compute the correlation of the feature maps at different scales~\cite{lin2017feature}: the features $\bm{H}_{rR}$ and $\bm{H}_{rF}$ are first downsampled a factor of 2 $k$ times, resulting two groups of feature maps $\{\bm{H}_{rR}^0, \bm{H}_{rR}^{1}, ... \bm{H}_{rR}^{k},\}$ and $\{\bm{H}_{rF}^0, \bm{H}_{rF}^{1}, ..., \bm{H}_{rF}^{k},\}$. For each scale $k$, we compute the Pearson correlation coefficient on each spatial position $(i,j)$ as:
\begin{equation}
    \bm{C}^{k, (i,j)} = \frac{\bm{H}_{rR}^{k,(i,j)} * \bm{H}_{rF}^{k, (i,j)}}{\Vert \bm{H}_{rR}^{k, (i,j)} \Vert ^2 \cdot \Vert \bm{H}_{rF}^{k, (i,j)} \Vert ^2} ,  \quad  C^k \in \mathbb{R}^{\frac{w}{2^k}\times \frac{h}{2^k}}
\end{equation}
where $*$ indicate dot product. It is important that \modulecorr contains fewest number of parameters so that most of the representation is learned in the \modulesemantic, so we choose to use correlation instead of spatial attention~\cite{wang2017residual}. Finally, all the correlation maps $\{\bm{C}^0, \bm{C}^1, ..., \bm{C}^k\}$ are upsampled to match the same size $w,h$ as $\bm{F}_{rR}$ and summed together to form a consensus map $\bm{C}$. 

The consensus map $\bm{C}$ is then used as a spatial weight map for the weighted average of feature maps $\bm{F}_{rR}$ and $\bm{F}_{rF}$. For generating more robust consensus map, we add a residual connection following~\cite{wang2017residual}, forming feature vectors $\bm{f}_{rR}$ and $\bm{f}_{rF}$. 

\subsection{Late fusion and adversarial training}
After processed by \modulesemantic and \modulecorr, for each modality a refined feature $\bm{f}_{rR}$, $\bm{f}_{rF}$ and $\bm{f}_{rA}$ is acquired, where the fusion of modalities can be adopted. For fusion these features, we concatenate $\bm{f}_{rR}$, $\bm{f}_{rF}$ to get a prediction score $s_1$, then fuse with the score $s_2$ generated by $\bm{f}_{rA}$ using weighted average. The weight for $s_1$ is 1 and for $s_2$ is 0.5.

Our full loss function is a combination of classification loss $\mathcal{L}_y$ and adversarial loss $\mathcal{L}_d$: 
\begin{equation}
    \mathcal{L} = \lambda_y \mathcal{L}_y + \lambda_d \mathcal{L}_d
\end{equation}

\section{Experiments}
We follow the experiment setup and use the train-val-test split as required by the challenge. 

\subsection{Feature extraction}
We use two backbones for feature extraction: pretrained I3D~\cite{carreira2017quo} and pretrained TBN~\cite{kazakos2019epic}, both of them are fine-tuned on the source training set. 
Additionally, object features extracted by Faster R-CNN object detector trained on EPIC-object-detection dataset is used as the object modality.
We also use hand-object bounding boxes to crop the input images, and extract features using TBN without further fine-tuning as the cropped-RGB modality and cropped-Flow modality. The bounding boxes are generated by a hand-object detector trained on 100 DOH dataset~\cite{Shan20}. We take the maximum of all detected boxes as the crop area. 

\subsection{Implementation Details}
The \modulesemantic processes the feature with dimension $c=1024$, and the ratio for gating bottleneck is $r=16$. We empirically choose $\lambda_y=1$ in all experiments, $\lambda_d=3$ for experiments with I3D backbone and $\lambda_d=1$ otherwise. For all experiments, we train the model 30 epochs on 4 NVIDIA-V100 GPUs. 

\subsection{Result}
Table~\ref{tab:epic100-test} demonstrates the recognition performance on target test set. Using RGB, Flow and Audio modalities and the same backbone TBN, our proposed method performs favorably against TA$^3$N~\cite{chen2019temporal} by 1.46\% in terms of the accuracy of action.

\begin{table}[h]
    \centering
    \resizebox{0.9\linewidth}{!}{
    \begin{tabular}{ccccccc}
    \toprule
    \multirow{2}{*}{Module}&
    \multicolumn{3}{c}{Top-1}&
    \multicolumn{3}{c}{Top-5}\cr
    \cmidrule(lr){2-4}
    \cmidrule(lr){5-7}
    &Verb&Noun&Action &Verb&Noun&Action\cr 
    \midrule
    TA$^3$N & 46.91 & 27.69 & 18.95 & 72.70 & 50.72 & 30.53 \cr 
    TA$^3$N+Ours & 49.99 & 30.45 & 20.41 & 77.97 & 54.58 & 35.20 \cr
    \bottomrule
    \end{tabular}}
    \vspace{0.3cm}
    \caption{Comparison of action recognition result on the target test set.}
    \label{tab:epic100-test}
    \vspace{-0.5cm}
\end{table}

\subsection{Model ensemble}
To take advantages of models trained with different inputs or different backbones, we explore model ensemble technique to fuse the following models:
\begin{itemize}
    \item Model A: taking RGB, Flow and Audio modalities as inputs, and using TBN  as the feature extraction backbone.
    \item Model B: compared with Model A, adding object features as an additional modality.
    \item Model C: taking cropped-RGB, cropping-Flow and Audio modalities as inputs, and using TBN as the feature extraction backbone.
    \item Model D: taking RGB and Flow modalities as inputs, and using I3D as the feature extraction backbone.
\end{itemize}
All of the four models use TA$^3$N+Ours as the domain adaptation module.

\begin{table}[h]
    \centering
    \resizebox{0.9\linewidth}{!}{
    \begin{tabular}{ccccccc}
    \toprule
    \multirow{2}{*}{Models}&
    \multicolumn{3}{c}{Top-1}&
    \multicolumn{3}{c}{Top-5}\cr
    \cmidrule(lr){2-4}
    \cmidrule(lr){5-7}
    &Verb&Noun&Action &Verb&Noun&Action\cr 
    \midrule
    A+B & 51.45 & 34.07 & 22.93 & 80.88 & 59.03 & 38.69 \cr
    A+B+C & 52.60 & 35.32 & 24.13 & 81.30 & 59.57 & 39.96 \cr
    A+B+C+D & 53.29 & 35.64 & 24.76 & 81.64 & 59.89 & 40.73 \cr
    \bottomrule
    \end{tabular}}
    \vspace{0.3cm}
    \caption{Model ensemble results on the target test set.}
    \label{tab:ensemble}
    \vspace{-0.5cm}
\end{table}

\section{Conclusion}
In this report, we introduce a novel Multi-Modal Mutual Enhancement module, which enables the mutual refinement between multiple modalities. The experimental result validates that our \model can significantly improve the domain adaptive action recognition performance. With model ensemble technique, we achieve competitive results on the leaderboard of the 2021 EPIC-KITCHENS-100 Unsupervised Domain Adaptation Challenge.

{\small
	\bibliographystyle{ieee_fullname}
	\bibliography{egbib}
}

\end{document}